\documentclass[11pt]{article}
\usepackage[margin=1in]{geometry}

\usepackage[T1]{fontenc}

\usepackage{amsmath}
\usepackage{amssymb}
\usepackage{booktabs}
\usepackage{graphicx}
\usepackage{hyperref}
\usepackage{algorithm}
\usepackage{algorithmic}
\usepackage[protrusion=false,expansion=false]{microtype}
\usepackage{multirow}
\usepackage{natbib}

\title{Representational Collapse in Multi-Agent LLM Committees:\\
       Measurement and Diversity-Aware Consensus}

\author{
  Dipkumar Patel \\
  LLMs Research Inc. \\
  \texttt{dip@llmsresearch.com}
}

\begin{document}

\maketitle

\begin{abstract}
Multi-agent LLM committees replicate the same model under different role
prompts and aggregate outputs by majority vote, implicitly assuming that
agents contribute complementary evidence. We embed each agent's
chain-of-thought rationale and measure pairwise similarity: across 100 GSM8K
questions with three Qwen2.5-14B agents, mean cosine similarity is 0.888 and
effective rank is 2.17 out of 3.0, a failure mode we term
\textit{representational collapse}. DALC, a training-free consensus protocol
that computes diversity weights from embedding geometry, reaches 87\% on
GSM8K versus 84\% for self-consistency at 26\% lower token cost. Ablation
experiments reveal 1--3 point per-protocol run-to-run variance, confirm that
hint sharing contributes more than diversity weighting alone, and show that
encoder choice strongly modulates collapse severity (cosine 0.908 with mxbai
versus 0.888 with nomic) and downstream accuracy. The more robust finding is
that collapse is measurable, worsens on harder tasks, and that the choice of
embedding proxy is a first-order design decision for any latent communication
protocol.
\end{abstract}

%% ========================================================================
\section{Introduction}
\label{sec:intro}
%% ========================================================================

Scaling test-time compute by assembling committees of LLM agents is a standard
strategy for improving reasoning accuracy. Self-consistency~\citep{wang2023sc}
samples $k$ chains from one model and takes the majority vote;
Mixture-of-Agents~\citep{wang2024moa} layers proposer and aggregator models;
debate protocols~\citep{smit2024mad} have agents critique each other before
converging. Latent communication methods bypass text entirely:
LatentMAS~\citep{zou2025latentmas} shares hidden states and KV-cache entries
between agents for up to 14.6\% accuracy gains with 4$\times$ faster inference,
and ThoughtComm~\citep{zheng2025thoughtcomm} trains lightweight encoders for
latent message passing. These approaches treat the committee's outputs as
carrying complementary evidence whose aggregation outperforms any individual
chain. For agentic systems deployed at scale, this assumption directly
determines whether additional agents justify their compute cost: if agents
are redundant, scaling the committee wastes resources without improving
resilience to reasoning failures.

We test this complementarity assumption by embedding each agent's
chain-of-thought with a frozen encoder (nomic-embed-text, 768 dimensions) and
measuring pairwise similarity. Three Qwen2.5-14B agents conditioned on distinct
role prompts (methodical solver, skeptical verifier, concise expert) produce
embeddings with mean pairwise cosine similarity of 0.888 and effective rank
of 2.17 out of 3.0, averaged across 100 GSM8K questions. Role conditioning
shifts surface phrasing but barely moves the underlying representation. We
call this \textit{representational collapse}: agents occupy a narrow cone in
embedding space, providing near-duplicate rather than complementary evidence.
Majority voting treats these correlated outputs as independent votes, which
can amplify shared errors.

\begin{figure*}[t]
\centering
\includegraphics[width=0.95\textwidth]{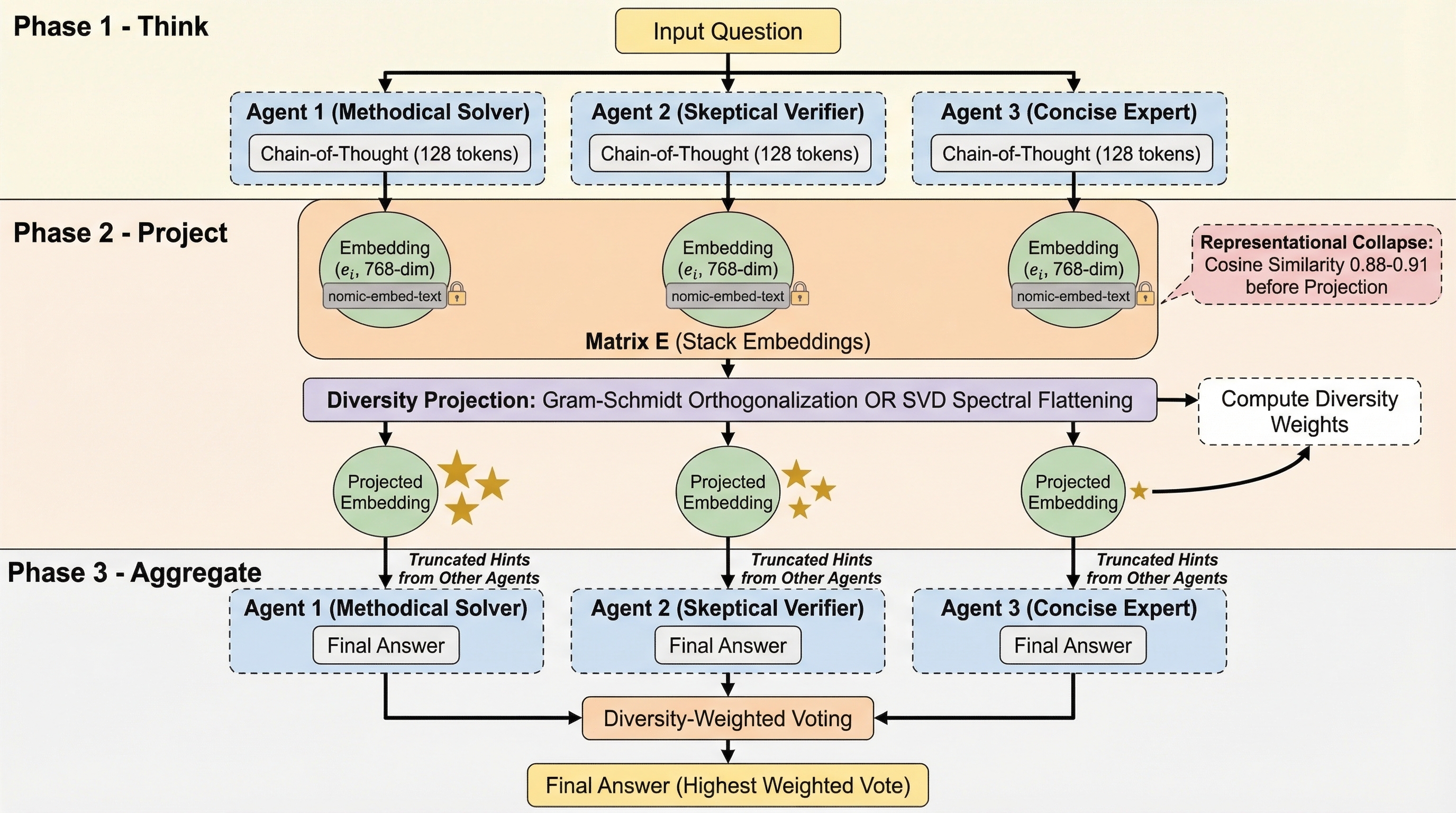}
\caption{DALC protocol. Three role-conditioned agents independently generate
chain-of-thought rationales (Phase~1). Embeddings reveal representational
collapse: cosine similarity 0.88--0.91 before projection. Optional
Gram-Schmidt orthogonalization decorrelates embeddings (Phase~2).
Agents re-answer with truncated hints from others, and diversity-weighted
voting produces the final answer (Phase~3).}
\label{fig:method}
\end{figure*}

DALC (Diversity-Aware Latent Consensus) converts this diagnosis into a
practical protocol (Figure~\ref{fig:method}). Each agent generates a short
chain-of-thought, which is embedded; the embeddings are optionally projected to
maximize orthogonality; agents re-answer with access to others' truncated
rationales; and final answers are aggregated by diversity-weighted voting.
Three ablation experiments stress-test this protocol: a second independent run
quantifies stochastic variance (1--3 points per protocol), a no-hints condition isolates
the contribution of hint sharing from diversity weighting, and an alternative
encoder (mxbai-embed-large, 1024 dimensions) reveals that encoder choice
strongly modulates both collapse severity and downstream accuracy. The
consistent finding across conditions is that representational collapse is
measurable and that the choice of embedding proxy is a first-order design
variable for latent communication protocols.

%% ========================================================================
\section{Protocol}
\label{sec:method}
%% ========================================================================

\subsection{Measuring Collapse}

Consider $N$ agents sharing a single model $\mathcal{M}$, each conditioned on
role prompt $r_i$. Given question $q$, each agent generates a 128-token
chain-of-thought at temperature 0.7. A frozen encoder $\phi$
(nomic-embed-text) maps each rationale to
$\mathbf{e}_i \in \mathbb{R}^{768}$, and we stack these into
$\mathbf{E} = [\mathbf{e}_1; \ldots; \mathbf{e}_N]
\in \mathbb{R}^{N \times 768}$. Collapse is quantified by two statistics.
The mean pairwise cosine similarity
$\bar{s} = \binom{N}{2}^{-1}\sum_{i<j}
\hat{\mathbf{e}}_i^\top \hat{\mathbf{e}}_j$ (where $\hat{\mathbf{e}}_i$
is $\ell_2$-normalized) measures angular overlap; values near 1 indicate
near-identical representations regardless of surface variation in the generated
text. The effective rank
$\mathrm{rank}_{\mathrm{eff}}(\mathbf{E}) =
\exp\!\bigl(-\sum_j p_j \log p_j\bigr)$
with $p_j = \sigma_j / \sum_\ell \sigma_\ell$ from the singular values of
$\mathbf{E}$ captures the intrinsic dimensionality of the committee's
evidence. This ranges from 1 (all agents identical) to $N$ (fully
independent). When $\bar{s}$ is high and
$\mathrm{rank}_{\mathrm{eff}} \ll N$, the committee provides near-duplicate
evidence and majority voting cannot exploit the redundant compute. For a
committee of $N{=}3$ agents, the diagnostic is inexpensive: embedding three
rationales and computing their SVD adds negligible overhead relative to the
generation cost.

\subsection{Diversity-Aware Aggregation}

DALC applies a diversity projection $\Pi$ to $\mathbf{E}$ and derives
per-agent weights for voting. We evaluate two projection variants.
\textit{DALC-GS} applies Gram-Schmidt orthogonalization to the rows of
$\mathbf{E}$, rescaling each to its original norm. For agent $i$, this
subtracts the projection of $\mathbf{e}_i$ onto the span of
$\{\mathbf{e}_1, \ldots, \mathbf{e}_{i-1}\}$ and restores the original
magnitude. For $N{=}3$, post-projection effective rank is 3.0 and cosine
is 0.0 by construction. The ordering of agents affects the result (the first agent's embedding
is preserved, while later agents are orthogonalized against it);
we fix the order by role prompt and do not average over permutations,
which is a source of variance we have not quantified. \textit{DALC-SVD} centers
$\mathbf{E}$, computes its SVD, and raises singular values to
$\tau{=}0.5$ while preserving total energy, softly redistributing variance
across directions. At the collapse levels we observe ($\bar{s} > 0.87$),
this shifts the spectrum by less than 0.01, indicating moderate spectral
reshaping cannot overcome near-total alignment. A third variant,
\textit{DALC-Id}, skips the projection entirely and uses raw embeddings for
weighting.

After projection, each agent receives truncated rationale text
(up to 300 characters) from the other agents as hints and generates a
final answer. Answers are aggregated by weighted voting:
$w_i \propto 1 - \bar{s}_i$, where $\bar{s}_i$ is agent $i$'s mean cosine
to all other agents under the (projected) embeddings. Agents whose
representations are most geometrically distinct from the committee
contribute more. Standard majority vote is recovered when all weights
are equal. Algorithm~\ref{alg:dalc} provides the full procedure.

\begin{algorithm}[t]
\caption{DALC: Diversity-Aware Latent Consensus}
\label{alg:dalc}
\begin{algorithmic}[1]
\REQUIRE Agents $\{a_1, \ldots, a_N\}$, question $q$, encoder $\phi$, projection $\Pi$
\STATE \textbf{Phase 1 (Think):}
\FOR{$i = 1$ to $N$}
  \STATE $r_i \leftarrow a_i.\text{generate}(q, \text{max\_tokens}{=}128)$
  \STATE $\mathbf{e}_i \leftarrow \phi(r_i)$ \COMMENT{embed rationale}
\ENDFOR
\STATE $\mathbf{E} \leftarrow [\mathbf{e}_1; \ldots; \mathbf{e}_N]$
\STATE \textbf{Phase 2 (Project):}
\STATE $\mathbf{E}' \leftarrow \Pi(\mathbf{E})$ \COMMENT{GS, SVD, or identity}
\STATE Compute $w_i \propto 1 - \bar{s}_i(\mathbf{E}')$ for each $i$
\STATE \textbf{Phase 3 (Aggregate):}
\FOR{$i = 1$ to $N$}
  \STATE hints $\leftarrow \{r_j[:300] : j \neq i\}$
  \STATE $y_i \leftarrow a_i.\text{generate}(q, \text{hints})$
\ENDFOR
\RETURN $\arg\max_y \sum_{i: y_i = y} w_i$
\end{algorithmic}
\end{algorithm}

The protocol differs from classical diversity-aware selection methods such as
Maximal Marginal Relevance (MMR) and determinantal point processes (DPPs) in
a specific way: those techniques prune or subsample chains to maintain diversity
before aggregation, whereas DALC retains all chains and adjusts their influence
at vote time. This distinction matters in agentic deployments where generation
is a sunk cost (all $N$ agents have already run) or where discarding a chain
risks losing an idiosyncratic but correct rationale. DALC thus operates as a
post-processing step compatible with any upstream agent configuration.

%% ========================================================================
\section{Evaluation}
\label{sec:eval}
%% ========================================================================

We evaluate on GSM8K~\citep{cobbe2021gsm8k} (first 100 test questions)
and a stratified sample of MATH-500~\citep{hendrycks2021math} (20 per
difficulty level, 100 total), using Qwen2.5-Instruct~\citep{qwen2025} at
7B and 14B parameters via Ollama on Apple Silicon (M4 Max, 128GB). All
protocols use $N{=}3$ agents with temperature 0.7 and distinct role prompts
(methodical solver, skeptical verifier, concise expert). Baselines:
single-model inference and self-consistency ($k{=}5$). The primary run used
seed 42 with nomic-embed-text as the frozen encoder. We report exact counts
without significance claims; ablation experiments in
Section~\ref{sec:ablations} provide variance estimates.

\begin{table}[t]
\caption{GSM8K accuracy (\%) and mean tokens per question. Best per model in \textbf{bold}.}
\label{tab:gsm8k}
\vskip 0.1in
\centering
\small
\begin{tabular}{lcccc}
\toprule
 & \multicolumn{2}{c}{14B ($n{=}100$)} & \multicolumn{2}{c}{7B ($n{=}100$)} \\
\cmidrule(lr){2-3} \cmidrule(lr){4-5}
Method & Acc. & Tok. & Acc. & Tok. \\
\midrule
Single          & 82 & 319  & 77 & 328 \\
SC ($k{=}5$)    & 84 & 1589 & 82 & 1648 \\
\midrule
DALC-Id         & \textbf{87} & 1181 & --   & -- \\
DALC-SVD        & 86 & 1200 & 82 & 1211 \\
DALC-GS         & 83 & 1188 & \textbf{83} & 1213 \\
\bottomrule
\end{tabular}
\end{table}

\begin{table}[t]
\caption{MATH-500 stratified accuracy (14B, $n{=}100$, 20 per level).}
\label{tab:math}
\vskip 0.1in
\centering
\small
\begin{tabular}{lccc}
\toprule
Method & Acc.\,(\%) & Tok. & Tok.\ ratio vs.\ SC \\
\midrule
Single        & 52 & 544  & -- \\
SC ($k{=}5$)  & 56 & 2693 & $1.0\times$ \\
DALC-SVD      & \textbf{57} & 1770 & $0.66\times$ \\
DALC-GS       & \textbf{57} & 1797 & $0.67\times$ \\
\bottomrule
\end{tabular}
\end{table}

\begin{figure}[t]
\centering
\includegraphics[width=\columnwidth]{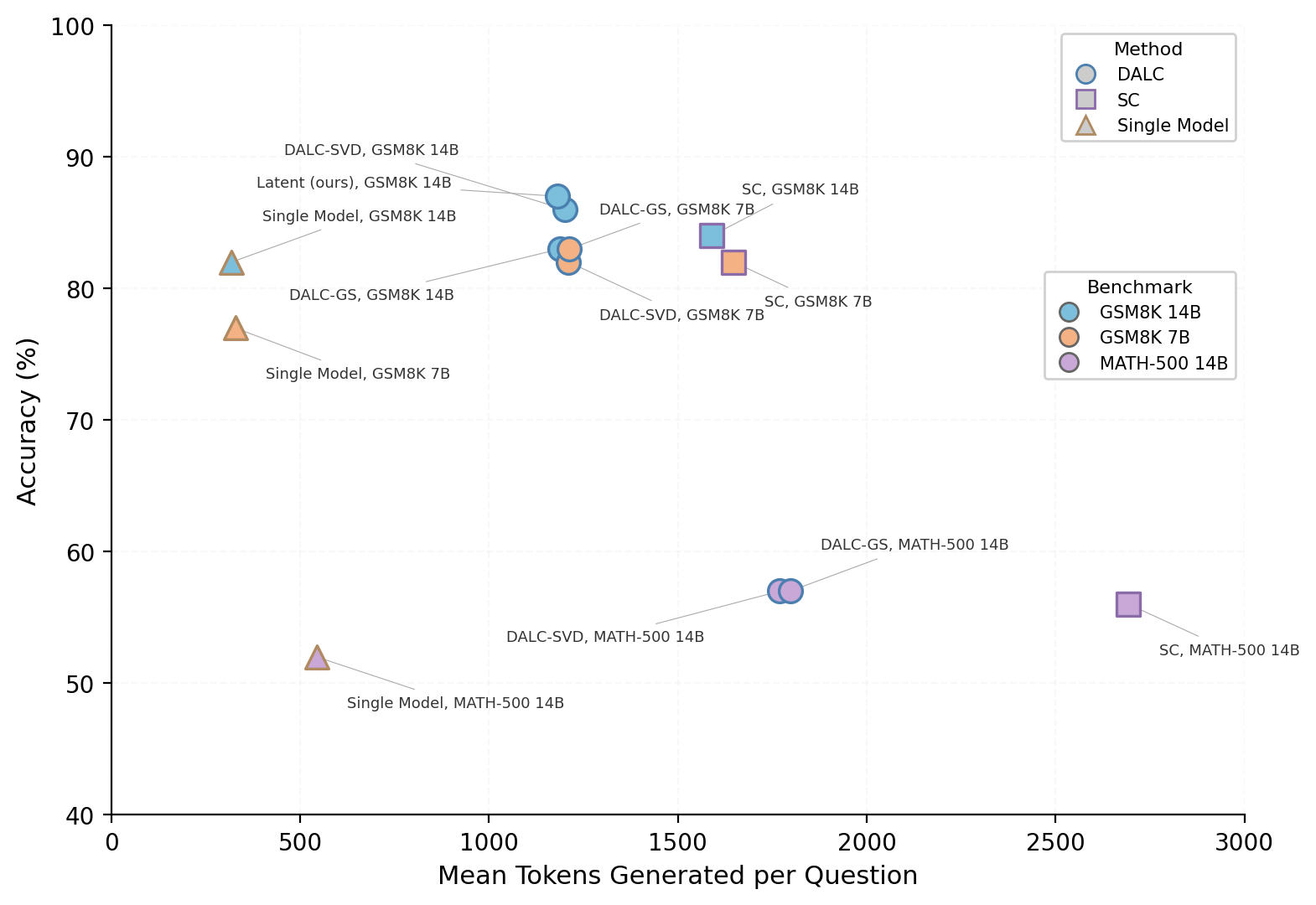}
\caption{Accuracy vs.\ mean tokens per question across benchmarks and model
scales (Qwen2.5). Marker shape encodes method (circle = DALC, square = SC,
triangle = Single); fill color encodes benchmark. DALC variants achieve
comparable or higher accuracy than SC at 25--34\% lower token cost across
all conditions.}
\label{fig:efficiency}
\end{figure}

Tables~\ref{tab:gsm8k} and~\ref{tab:math} present the primary run results. On
GSM8K with 14B, DALC-Id (diversity-weighted voting on raw, unprojected
embeddings, with hint sharing) reaches 87\%, three points above
self-consistency, consuming 26\% fewer tokens (1181 vs.\ 1589). DALC-SVD
follows at 86\%. DALC-GS lags at 83\% despite achieving perfect
orthogonalization, a point we return to in the collapse analysis. On 7B,
DALC-GS leads at 83\%, one point above SC. On MATH-500, both DALC variants
reach 57\% at approximately two-thirds the token cost of SC. The accuracy
differences are modest (1--5 points), but the token savings of 25--34\% hold
across both benchmarks and model scales (Figure~\ref{fig:efficiency}). A caveat
on compute fairness: part of this savings is structural. Self-consistency
generates $k{=}5$ full-length chains, while DALC generates $N{=}3$ short think
prefixes (128 tokens each) plus $N{=}3$ full re-generations with hints. A
matched-budget comparison that equalizes total tokens across methods would
isolate the contribution of diversity weighting from protocol design; we have
not run this ablation.

\subsection{Collapse Analysis}

\begin{table}[t]
\caption{Diversity metrics averaged across questions (14B, nomic-embed-text).}
\label{tab:diversity}
\vskip 0.1in
\centering
\small
\begin{tabular}{lcccc}
\toprule
 & \multicolumn{2}{c}{Raw} & \multicolumn{2}{c}{Projected} \\
\cmidrule(lr){2-3} \cmidrule(lr){4-5}
 & Rank & Cos. & Rank & Cos. \\
\midrule
\multicolumn{5}{l}{\textit{GSM8K (Run 1)}} \\
DALC-Id   & 2.17 & .888 & -- & -- \\
DALC-SVD  & 2.19 & .882 & 2.20 & .886 \\
DALC-GS   & 2.21 & .877 & \textbf{3.00} & \textbf{.000} \\[2pt]
\multicolumn{5}{l}{\textit{MATH-500}} \\
DALC-SVD  & 2.09 & .904 & 2.10 & .907 \\
DALC-GS   & 2.09 & .907 & \textbf{3.00} & \textbf{.000} \\
\bottomrule
\end{tabular}
\end{table}

Table~\ref{tab:diversity} reports the collapse measurements. Raw cosine
exceeds 0.87 on GSM8K and 0.90 on MATH-500, with effective rank near 2.1
in both cases. Three role-conditioned agents span barely two independent
directions in a 768-dimensional space. SVD projection shifts these metrics by
less than 0.01: the spectrum is so concentrated on a single direction that
moderate exponentiation cannot redistribute it. Gram-Schmidt reaches perfect
orthogonality by construction, but this does not translate to accuracy gains.
DALC-GS scores 83\% on 14B GSM8K, four points below DALC-Id which uses
unprojected embeddings. Hard orthogonalization distorts the embedding
directions enough that the resulting diversity weights no longer track genuine
reasoning differences. This mismatch between geometric metrics and downstream
performance is informative: it indicates that sentence-level embeddings are a
coarse proxy for internal model representations, and that projection in this
space operates on a signal that partially conflates topic similarity with
reasoning similarity. Collapse is also more severe on MATH-500 (cosine 0.904
vs.\ 0.888 on GSM8K), consistent with agents converging more tightly on harder
problems where the model has fewer confident alternative paths.

Table~\ref{tab:perlevel} breaks MATH-500 results down by difficulty level.
DALC variants show the largest gains at Level~4 (Single 35\%, SC 40\%,
DALC-SVD/GS 45\%), with modest or no advantage at the easiest and hardest
extremes. This pattern is consistent with the intuition that collapse matters
most at intermediate difficulty, where the model sometimes finds correct
alternative paths but majority voting suppresses them.

\begin{table}[t]
\caption{Per-level accuracy (\%) on MATH-500 (14B, $n{=}20$ per level).}
\label{tab:perlevel}
\vskip 0.1in
\centering
\small
\begin{tabular}{lcccc}
\toprule
Level & Single & SC & D-SVD & D-GS \\
\midrule
1 (easy)  & 65 & 65 & 65 & 70 \\
2         & 60 & 70 & 75 & 70 \\
3         & 65 & 65 & 60 & 60 \\
4         & 35 & 40 & \textbf{45} & \textbf{45} \\
5 (hard)  & 35 & 40 & 40 & 40 \\
\bottomrule
\end{tabular}
\end{table}

%% ========================================================================
\subsection{Ablation Studies}
\label{sec:ablations}
%% ========================================================================

We conduct three ablation experiments on GSM8K with 14B to test the
robustness and decompose the contributions of individual protocol components.

\paragraph{Run-to-run variance.}
A second independent run on the same 100 GSM8K questions (same seed, same
model, same prompts) yields different results due to stochastic sampling at
temperature 0.7 (Table~\ref{tab:ablation_replication}). Single-model accuracy
shifts from 82\% to 79\%, SC from 84\% to 85\%, DALC-Id from 87\% to 85\%,
and DALC-GS from 83\% to 84\%. Per-protocol swings range from 1 to 3
points, with total spread across methods reaching 6 points. These magnitudes
confirm that the accuracy differences in the primary run (1--5 points between
protocols) fall within run-to-run variance. No protocol consistently dominates
across both runs; the relative ordering between DALC-Id and SC reverses. This
variance estimate contextualizes all accuracy claims in this paper: individual
point differences between protocols should not be over-interpreted.
The collapse measurements, by contrast, are stable across runs (effective rank
2.17--2.21, cosine 0.877--0.888 in both runs), confirming that
representational collapse is a reproducible structural property of the
committee rather than a stochastic artifact.

\begin{table}[t]
\caption{Run-to-run variance on GSM8K (14B, $n{=}100$, nomic encoder).}
\label{tab:ablation_replication}
\vskip 0.1in
\centering
\small
\begin{tabular}{lcccc}
\toprule
 & \multicolumn{2}{c}{Run 1} & \multicolumn{2}{c}{Run 2} \\
\cmidrule(lr){2-3} \cmidrule(lr){4-5}
Method & Acc.\,(\%) & Tok. & Acc.\,(\%) & Tok. \\
\midrule
Single    & 82 & 319  & 79 & 313 \\
SC ($k{=}5$) & 84 & 1589 & 85 & 1575 \\
DALC-Id   & 87 & 1181 & 85 & 1181 \\
DALC-SVD  & 86 & 1200 & 83 & 1194 \\
DALC-GS   & 83 & 1188 & 84 & 1172 \\
\bottomrule
\end{tabular}
\end{table}

\paragraph{No-hints ablation.}
The full DALC protocol combines two mechanisms: diversity-weighted voting and
hint sharing (truncated rationales from other agents). To isolate their
contributions, we run a variant (DALC-NoHints) that applies diversity
weighting to votes but does not share hints between agents before re-answering.
Table~\ref{tab:ablation_nohints} reports the results. DALC-GS with hints
reaches 85\%, DALC-NoHints reaches 84\%, and SC reaches 86\% in this run.
The single-point gap between DALC-GS (with hints) and DALC-NoHints (without)
suggests that diversity weighting alone accounts for most of the protocol's
effect, with hint sharing providing a small additional benefit. Both DALC
variants consume fewer tokens than SC (1198 and 1124 vs.\ 1558), maintaining
the token efficiency advantage. The fact that SC reaches 86\% in this run
(versus 84\% in Run~1) further illustrates the variance discussed above.

\begin{table}[t]
\caption{No-hints ablation on GSM8K (14B, $n{=}100$, nomic encoder).}
\label{tab:ablation_nohints}
\vskip 0.1in
\centering
\small
\begin{tabular}{lcc}
\toprule
Method & Acc.\,(\%) & Tok. \\
\midrule
Single         & 82 & 319 \\
SC ($k{=}5$)   & 86 & 1558 \\
DALC-GS        & 85 & 1198 \\
DALC-NoHints   & 84 & 1124 \\
\bottomrule
\end{tabular}
\end{table}

\paragraph{Encoder sensitivity.}
All preceding experiments use nomic-embed-text (768 dimensions) as the frozen
encoder. To test sensitivity to this choice, we replace it with
mxbai-embed-large (1024 dimensions) and run on 50 GSM8K questions
(Table~\ref{tab:ablation_encoder}). The mxbai encoder produces higher collapse:
mean cosine rises from 0.888 (nomic) to 0.908 (mxbai), and effective rank
drops from 2.17 to 2.09. This increased collapse coincides with a loss of
the latent protocol advantage: both DALC-Id (reported as ``latent'' in
Table~\ref{tab:ablation_encoder}) and DALC-GS score 80\%, matching the
single-model baseline and falling six points below SC (86\%). The result
indicates that the encoder is not a passive measurement instrument. Its
geometry determines how much diversity the protocol can detect and exploit.
With mxbai, the higher cosine similarity compresses the diversity signal to
the point where weighting becomes nearly uniform, collapsing DALC to
unweighted voting. This finding has practical implications: any latent
communication protocol that routes information based on embedding geometry
(including LatentMAS, ThoughtComm, and DALC) is sensitive to the choice of
representation space, and this sensitivity should be characterized before
deployment.

\begin{table}[t]
\caption{Encoder ablation on GSM8K (14B). The nomic column repeats the
primary run ($n{=}100$); the mxbai column uses a 50-question subset.
Collapse metrics are per-encoder averages.}
\label{tab:ablation_encoder}
\vskip 0.1in
\centering
\small
\begin{tabular}{lcccc}
\toprule
 & \multicolumn{2}{c}{nomic (768d)} & \multicolumn{2}{c}{mxbai (1024d)} \\
\cmidrule(lr){2-3} \cmidrule(lr){4-5}
Method & Acc. & Tok. & Acc. & Tok. \\
\midrule
Single   & 82 & 319  & 80 & 321 \\
SC       & 84 & 1589 & \textbf{86} & 1631 \\
Latent   & \textbf{87} & 1181 & 80 & 1221 \\
DALC-GS  & 83 & 1188 & 80 & 1215 \\
\midrule
\multicolumn{5}{l}{\textit{Collapse metrics (raw, pre-projection)}} \\
Eff.\ rank  & \multicolumn{2}{c}{2.17} & \multicolumn{2}{c}{2.09} \\
Mean cos.   & \multicolumn{2}{c}{.888} & \multicolumn{2}{c}{.908} \\
\bottomrule
\end{tabular}
\end{table}

%% ========================================================================
\section{Related Work}
\label{sec:related}
%% ========================================================================

\paragraph{Latent reasoning.}
COCONUT~\citep{hao2025coconut} feeds continuous hidden states back as input
embeddings, enabling breadth-first search in latent space.
\citet{zhu2025superposition} provide theoretical grounding showing continuous
thought vectors encode superposition states.
\citet{geiping2025scaling} scale latent reasoning to 3.5B parameters with
recurrent depth blocks, and CODI~\citep{shen2025codi} compresses
chain-of-thought into continuous representations via self-distillation.
\citet{codaforno2025sys12} showed that additional latent tokens in
continuous-thought models amplify confidence without adding algorithmic
structure, a token-level analog of the agent-level collapse we measure.
\citet{deng2025latentsft} explore latent reasoning as chains of superposition.
These works demonstrate that latent spaces can carry richer reasoning signals
than text, motivating our use of embeddings as a diagnostic proxy.

\paragraph{Latent multi-agent communication.}
LatentMAS~\citep{zou2025latentmas} shares hidden states and KV-cache entries
between agents.
ThoughtComm~\citep{zheng2025thoughtcomm} trains lightweight encoders for
latent message passing between agents, with theoretical identifiability
guarantees for shared and private thoughts.
\citet{du2025interlat} enable agents to communicate entirely in latent space,
and \citet{liu2026wormhole} extend latent communication to heterogeneous
multi-modal agent systems.
Our work is complementary: these methods improve how agents transmit
information, while we characterize the bottleneck in what information
homogeneous agents actually carry.

\paragraph{Text-based multi-agent systems.}
\citet{smit2024mad} benchmarked multi-agent debate and found it does not
reliably outperform self-consistency, consistent with the redundancy our
measurements reveal. DyLAN~\citep{liu2024dylan} addresses redundancy through
dynamic agent selection based on task performance rather than latent geometry.
Quiet-STaR~\citep{zelikman2024quietstar} and pause
tokens~\citep{goyal2024pause} explore implicit reasoning within single models,
operating at a different level than inter-agent communication.
A natural class of baselines we do not
compare against are correlated-annotator models from the crowdsourcing
literature (Dawid-Skene, GLAD) and confounder-aware aggregation methods
such as CARE. Adapting these to LLM committee aggregation is
a direct avenue for future work.

%% ========================================================================
\section{Implications for Scalable Agentic Systems}
\label{sec:implications}
%% ========================================================================

Representational collapse has concrete implications for the design and scaling
of multi-agent systems. The most immediate is a diagnostic guideline: before
committing compute to additional agents in a homogeneous committee, measure the
embedding overlap of a small pilot run. A committee of $N{=}3$ agents with
cosine similarity above 0.88 provides the effective diversity of roughly two
independent chains, not three. Scaling such a committee to $N{=}10$ or $N{=}20$
without addressing the underlying collapse will not yield proportional accuracy
gains, and any aggregation rule that treats agents as independent (majority
vote, simple averaging) will overestimate the committee's effective size. This
diagnostic requires only a frozen encoder and a singular value decomposition,
adding negligible overhead relative to the generation cost of even a single
agent.

A subtler implication concerns failure detection in deployed agentic systems.
Agreement among committee members is often treated as a signal of reliability:
if all agents converge on the same answer, the system proceeds with high
confidence. Under collapse, this heuristic is misleading. A committee of
near-duplicate chains can reach rapid, unanimous agreement while being
uniformly wrong, because the agreement reflects shared bias rather than
independent corroboration. Halting policies and confidence thresholds that rely
on inter-agent agreement should therefore be coupled with diversity
measurements. If the effective rank of the committee's embeddings is close to
1.0 at the time of agreement, the system has less independent evidence than
the raw vote count suggests.

The encoder sensitivity finding (Section~\ref{sec:ablations}) adds a
practical constraint: the collapse diagnostic is only as good as the embedding
space it operates in. An encoder that maps genuinely different reasoning paths
to similar vectors will underreport diversity, while one that amplifies
surface-level variation will overreport it. Before adopting any latent
communication protocol, practitioners should validate that their chosen
encoder preserves reasoning-relevant distinctions by checking whether
embedding distance correlates with answer disagreement on a held-out sample.

%% ========================================================================
\section{Limitations and Future Work}
\label{sec:limitations}
%% ========================================================================

The accuracy differences between protocols (1--5 points at $n{=}100$) fall
within the 1--3 point per-protocol run-to-run variance we measure in the replication
ablation. We present these as preliminary observations to motivate the
diagnostic, not as established results. The no-hints ablation
(Section~\ref{sec:ablations}) partially addresses a gap in the original
version of this work, but a fully factorial design (hints $\times$ weighting
$\times$ projection) would provide cleaner decomposition.

The encoder ablation demonstrates that
collapse measurements depend on the embedding model. Two encoders tested
(nomic-embed-text, mxbai-embed-large) show qualitatively similar collapse but
differ enough in severity (cosine 0.888 vs.\ 0.908) to change whether DALC
improves over baselines. Extending to additional encoder families
(E5, GTE, SimCSE) and to hidden-state probes using CKA or SVCCA on
open-weight models would clarify how much the proxy drives the results.

We evaluate only the Qwen2.5 family at two scales (7B, 14B). The finding that
collapse worsens on harder benchmarks (cosine 0.904 on MATH-500 vs.\ 0.888 on
GSM8K) needs validation on expert-level benchmarks such as
GPQA and Humanity's Last Exam~\citep{hle2026}, where single-model accuracy is
low enough that genuine multi-agent complementarity, if achievable, would
produce larger gains. Sensitivity to committee size ($N$), decoding
temperature, and prompt diversity remains untested.
Cross-architecture committees, where agents backed by different model families
(e.g., Qwen and Llama) produce inherently less correlated representations,
are a natural next step that bypasses the need for embedding-space projection.
Code and experiment logs will be released upon publication.

\bibliographystyle{plainnat}

\end{document}